\documentclass[sigconf,nonacm=true]{acmart}
\AtBeginDocument{%
  }

\usepackage{listings}

\usepackage{tcolorbox}
\usepackage{enumitem}

\newtcolorbox{promptbox}{
  colback=gray!5,
  colframe=black!40,
  boxrule=0.5pt,
  arc=2pt,
  left=6pt,
  right=6pt,
  top=6pt,
  bottom=6pt
}

\definecolor{codegreen}{rgb}{0,0.6,0}
\definecolor{codegray}{rgb}{0.5,0.5,0.5}
\definecolor{codepurple}{rgb}{0.58,0,0.82}
\definecolor{backcolour}{rgb}{0.97,0.97,0.95}
\definecolor{forestgreen}{rgb}{0.28,0.62,0.37}
\definecolor{codeblue}{rgb}{0,0.5,1}

\usepackage{multirow} 
\usepackage{makecell}

\usepackage{amsmath}

\lstdefinelanguage{markdown}{
    morekeywords={*, \#, \_, `},
    sensitive=false,
    morecomment=[l]{//},   
    morestring=[b]",        
    postbreak={},
breakindent=0pt,
breakautoindent=false,
}

\newcommand\pythonstyle{\lstset{
language=Python,
basicstyle=\ttm,
morekeywords={self},              
keywordstyle=\ttb\color{deepblue},
emph={MyClass,__init__},          
emphstyle=\ttb\color{deepred},    
stringstyle=\color{deepgreen},
frame=tb,                         
showstringspaces=false
}}

\lstnewenvironment{python}[1][]
{
\pythonstyle
\lstset{#1}
}
{}

\newcommand\pythoninline[1]{{\pythonstyle\lstinline!#1!}}

\lstdefinestyle{mystyle}{
    commentstyle=\color{codepurple},
    keywordstyle=\color{codepurple},
    numberstyle=\tiny\color{codegray},
    stringstyle=\color{blue},
    basicstyle=\ttfamily\small,
    breakatwhitespace=false,         
    breaklines=true,                 
    captionpos=b,                    
    keepspaces=true,                 
    showspaces=false,                
    showstringspaces=false,
    showtabs=false,                  
    tabsize=2
}

\usepackage{caption}
\usepackage{float}  

\usepackage{array} 
\usepackage{booktabs}
\usepackage[htt]{hyphenat}

\usepackage{cleveref}
\usepackage{enumitem}
\usepackage{algorithm}
\usepackage{algorithmicx}
\usepackage[noend]{algpseudocode}

\usepackage{xpatch}
\xpatchcmd\algorithmic{\leftmargin\labelwidth}{\leftmargin0.3\labelsep}{}{}
\algrenewcommand\alglinenumber[1]{\footnotesize #1}

\algrenewcommand\algorithmiccomment[1]{\hfill\textcolor{gray}{$\triangleright$ #1}}
\algrenewcommand\algorithmicindent{1em}

\algrenewcommand\algorithmicprocedure{\textbf{function}}

\usepackage{tikz}
\newcommand*\circled[1]{\protect\tikz[baseline=(char.base)]{\protect\node[shape=circle,fill=black,draw,inner sep=0.6pt] (char) {\textcolor{white}{\footnotesize \textbf{#1}}};}}

\definecolor{eidSafe}{HTML}{4C72B0}
\definecolor{eidUnsafe}{HTML}{DD8452}
\definecolor{eidMissing}{HTML}{CFCFCF}

\usepackage{pifont}
\newcommand{\xmark}{\ding{55}} 
\newcommand{\cmark}{\ding{51}} 

\usepackage[htt]{hyphenat}

\usepackage{multirow}   
\usepackage{booktabs}   

\usepackage{wrapfig}
\usepackage{subfigure}

\newcommand{\header}[1]{\vspace{1mm}\noindent\textbf{#1}.}

\newcommand{\headerl}[1]{\vspace{1mm}\noindent\textit{#1}.}

\newcommand{\system}{\texttt{PrismaDV}}
\newcommand{\icdbench}{\texttt{ICDBench}}
\newcommand{\eidbench}{\texttt{EIDBench}}

\begin{document}

\title{PrismaDV: Automated Task-Aware Data Unit Test Generation}

\author{Hao Chen}
\orcid{0009-0004-1887-6630}
\affiliation{%
  \institution{BIFOLD \& TU Berlin}
  \country{}
}
\email{hao.chen@tu-berlin.de}

\author{Arnab Phani}
\orcid{0009-0001-2935-0608}
\affiliation{%
  \institution{BIFOLD \& TU Berlin}
  \country{}
}
\email{arnab.phani@tu-berlin.de}

\author{Sebastian Schelter}
\orcid{0000-0003-4722-5840}
\affiliation{%
  \institution{BIFOLD \& TU Berlin}
  \country{}
}
\email{schelter@tu-berlin.de}

\renewcommand{\shortauthors}{Hao Chen, Arnab Phani, \& Sebastian Schelter}

\begin{abstract}
Data is a central resource for modern enterprises, and data validation is essential for ensuring the reliability of downstream applications. However, existing automated data unit testing frameworks are largely task-agnostic: they validate datasets without considering the semantics and requirements of the code that consumes the data.
We present \system{}, a compound AI system that analyzes downstream task code together with dataset profiles to identify data access patterns, infer implicit data assumptions, and generate task-aware executable data unit tests. To further adapt the data unit tests over time to specific datasets and downstream tasks, we propose ``Selective Informative Feedback for Task Adaptation'' (SIFTA), a prompt-optimization framework that leverages the scarce outcomes from the execution of data unit tests and downstream tasks. We evaluate \system{} on two new benchmarks spanning 60 tasks across five datasets, where it consistently outperforms both task-agnostic and task-aware baselines in generating unit tests that reflect the end-to-end impact of data errors. Furthermore, we show that with SIFTA, we can automatically learn prompts for \system{}'s modules that outperform prompts written by hand or generated from a generic prompt optimizer. We publicly release our benchmarks and prototype implementation.\end{abstract}

\maketitle

\section{Introduction}

Data is a central resource for modern enterprises and institutions, and data issues, such as missing or incorrect information~\cite{abedjan2016detecting,yan2020scoded,11107465}, can seriously impact their operations. Data errors propagating through data systems lead to serious impact in production, such as outages of mobile apps~\cite{facebookios}, bank customers losing access to their accounts~\cite{tsboutage}, outages of flights in the US~\cite{cnnfaa}, and the loss of medical records~\cite{ukcorona}. Furthermore, data errors are one of the major reasons for the silent performance degradation of deployed ML models~\cite{polyzotis2019data,schelter2015challenges,nigenda2022amazon}. A reason for this is that many organizations have adopted a ``collect first, analyze later'' workflow~\cite{DataSchoolHellersteinEp1_2024}, relying on the schema-on-read interpretation of data in downstream applications. Therefore, corrupted data often propagates unnoticed until it causes failures in production.

\header{Current landscape of data unit testing frameworks} As a consequence, data unit testing frameworks such as TensorFlow Data Validation~\cite{polyzotis2019data}, Amazon's Deequ~\cite{schelter2018automating,nigenda2022amazon,awsglue}, and Great Expectations~\cite{greatexpectations} have become widely used in industry in recent years. These frameworks generate data unit tests: declarative data constraints, often expressed in an easy-to-use DSL, such as null-value checks, completeness and uniqueness constraints, value-range checks, and distributional sanity checks. These constraints are inferred by profiling a data sample and subsequently applying heuristics, against which to validate unseen data. Major cloud providers offer data unit testing as part of their data infrastructure: Amazon's AWS Glue Data quality service~\cite{awsglue} defines a domain-specific language to enable non-coders to define data unit tests with Deequ, Databricks allows its users to annotate data pipelines with ``pipeline expectations''~\cite{databricksexpectations} for data unit testing, Google Dataplex~\cite{googledataplex} enables customers to choose from auto-suggested data quality rules for catching data anomalies in their data pipelines, and GXCloud recently announced an AI-driven constraint suggestion feature~\cite{expectai}.

\header{Shortcomings of current approaches} Despite their popularity, existing frameworks suffer from several shortcomings: $(i)$~authoring and maintaining data unit tests remains tedious and error-prone, since data engineers often write checks one column at a time, which does not scale to wide production tables with hundreds of columns~\cite{song2021auto}; as a result, validation coverage is typically partial and focuses on a subset of columns only. Frameworks such as Deequ and TFDV alleviate this burden by automatically suggesting constraints from sample data via data profiling, but this automation introduces additional challenges: $(ii)$~heuristically suggested constraints are often either too strict or too general; overly strict tests produce false alarms, leading to alert fatigue and costly on-call triage, while overly general tests tend to miss domain-specific data errors and can lead to production incidents that require substantial manual intervention. Furthermore, $(iii)$~designing effective data unit tests still requires manual post-editing by a data engineer with domain knowledge, and $(iv)$~over time, data unit tests typically evolve in a reactive way only: data problems become apparent in production systems, are manually fixed and tests are extended with additional constraints to prevent recurrence. This reactive cycle imposes a recurring tax on data teams: scarce engineering time is diverted to debugging, test maintenance, and incident response, slowing down feature and model iteration. Researchers proposed several extensions to address these shortcomings in recent years, which either leverage statistics from historical executions~\cite{redyuk2021automating,shankar2023automatic,tu2023auto,huang2018auto,song2021auto,dong2025automated} or require a human in the loop~\cite{raha,heidari2019holodetect,huang2024cocoon} to label data examples, neither of which directly leverages the downstream task code that actually consumes the data.

\header{Task-aware data unit test generation} We argue that a major limitation of current approaches is that they {\em rely on observed data only and ignore the characteristics of the downstream tasks that consume the data to validate}. This leads to several missed opportunities to improve data unit tests and address some of the outlined shortcomings. First, certain downstream tasks might only access parts of the data, especially for large denormalized datasets common in enterprise data lakes, which means that data unit tests for these tasks should focus on the subset of accessed columns only. Second, the code of downstream tasks is often written by experienced data engineers, with implicit domain knowledge about the data ``baked in'', which may be helpful to extract into a data unit test. Some downstream tasks like ML training tasks might even be naturally robust against certain types of noise in data, which a data unit test could account for. We illustrate these limitations with a running example in \Cref{sec:problem}.

A way forward is to improve the automated generation of data unit tests by {\em specializing them to the downstream tasks} for which they are deployed. However, this specialization is inherently difficult as it requires an ``understanding'' of downstream task code. In practice, production data pipelines support a diverse set of downstream tasks, ranging from recurring BI/ETL processing to web applications to display and edit data, to feature engineering and ML training or inference, and these tasks often encode different semantics and assumptions about the same data. Even seemingly simple problems like identifying which columns a piece of code accesses are challenging and typically handled via static code analysis with hand-curated knowledge bases~\cite{namaki2020vamsa}. Approaches like fuzzing-based testing~\cite{polyzotis2019data} are also difficult to apply in practice, as they assume that one can generate synthetic input data and repeatedly execute the downstream tasks in a ``sandbox mode''. In many industry settings, repeated execution is impractical because tasks have external side effects, for example materializing intermediate results, triggering actions in other systems, or customer interactions such as sending notification emails.

\header{Overview and contributions}
We propose to take downstream tasks into account for automated data unit test generation. To this end, we introduce \system{}, a task-aware data validation system that proactively~\cite{zeighami2025llm} generates specialized data unit tests for individual downstream tasks by jointly analyzing dataset profiles and task code. We motivate this direction with a running example~(\Cref{sec:problem}). We then describe the design of \system{}, a compound AI system~\cite{11108141}, which decomposes task-aware data unit test generation into multiple steps (data profiling, column access detection, assumption inference, and constraint code generation). We discuss these steps and how we leverage the code understanding~\cite{lester2025} and code synthesis capabilities of LLMs~\cite{le2025graph,huang2024cocoon} in \Cref{sec:approach}. To improve validation quality over time for a particular dataset, we propose ``Selective Informative Feedback for Task Adaptation'' (SIFTA), a lightweight prompt-optimization approach that leverages the scarce outcomes from the execution of data unit tests and downstream tasks as supervision signal (\Cref{sec:approach_optimization}). SIFTA identifies informative constraint failures via ``failure precision'' (the fraction of constraint failures that coincide with task failures), and backtraces these failures to the underlying data assumptions in code as input to an optimizer. Finally, we introduce two complementary benchmarks in \Cref{sec:benchmarks}: \icdbench{}, a hand-crafted benchmark for constraint discovery from data--code pairs, and \eidbench{}, an end-to-end benchmark with 60 downstream tasks across five public datasets. In summary, we provide the following contributions.
\begin{itemize}[leftmargin=*]
  \item We introduce the problem of task-aware data unit test generation~(\Cref{sec:problem}).
  \item We present \system{}, a compound AI system that analyzes downstream task code together with dataset profiles to identify data access patterns, infer implicit data assumptions, and generate task-aware executable data unit tests~(\Cref{sec:approach}).
  \item We propose SIFTA, a prompt optimization procedure for \system{} that leverages the scarce outcomes from the execution of data unit tests and downstream tasks, together with structured backtraces from constraints to assumptions and code~(\Cref{sec:approach_optimization}).
  \item We design two novel benchmarks for evaluating task-aware data unit test generation: \icdbench{} for individual constraint discovery from data--code pairs (63 cases with ground-truth constraints) and \eidbench{} for end-to-end error impact detection with five datasets, 60 tasks, and 25 error cases per dataset~(\Cref{sec:benchmarks}).
  \item We conduct an extensive experimental evaluation showing that \system{} outperforms strong baselines by more than 20 points in F1 score on \icdbench{}, more than 26 points in F1 score on \eidbench{}, and that SIFTA outperforms a general prompt optimizer~(\Cref{sec:experiments}).
  \item We make our code and benchmark available under an open license at \textcolor{blue}{\url{https://github.com/deem-data/PrismaDV}}.
\end{itemize}

\section{Background}

We briefly introduce the required background on data unit tests. Data unit tests are typically deployed as part of data pipelines which move data between different systems and applications~\cite{schelter2018automating,polyzotis2019data,databricksexpectations,awsglue,nigenda2022amazon}. The goal of a data unit test is to flag potentially erroneous data early to allow engineers to intervene before the data already caused issues in downstream applications. Data unit tests are crucial for the data operations in large organizations, where data updates are regularly produced and consumed by hundreds of downstream applications.

Formally, a data unit test $C = \{c_1, \dots, c_n\}$ consists of a set of constraints $\{c_1, \dots, c_n\}$. Each constraint $c_i$ is a variant of a primitive aggregation constraint~\cite{ross1998foundations}. In data unit testing libraries such as \texttt{pydeequ}~\cite{pydeequ}, constraints are declared as follows \texttt{hasCompleteness("colA", lambda x: x >= 0.99).where("colB > 10")}. This constraint states: {\em The column ``colA'' must have at least 99\% non-null values in rows where the corresponding value of ``colB'' is larger than ten}. Data unit tests are explicitly designed to rely on efficiently computable aggregates, since these tests must be runnable on datasets with billions of tuples~\cite{schelter2018automating}. For that reason, they often use approximations for expensive statistics, e.g., hyperloglog sketches~\cite{harmouch2017cardinality} for cardinality estimates or KLL sketches~\cite{karnin2016optimal} for approximating percentiles. Evaluating the data unit test $C$ on a dataset $D$ requires the evaluation of each constraint $c_i \in C$. The data unit test rejects $D$ if there exists a constraint which is not satisfied on $D$.

Designing data unit tests is challenging since it requires intricate knowledge about invariants of the data and the domain in which it is used. Furthermore, there is a tension between overly strict constraints, which may produce many false alarms and too general constraints, which may not be helpful in identifying issues in the data. Popular libraries like Deequ~\cite{deequdoc} and Tensorflow Data Validation~\cite{tfdvnotebook} offer automated ways to suggest constraints based on data profiling, which must typically be post-edited by data engineers.


\begin{figure*}[t!]
\centering
\includegraphics[width=\textwidth]{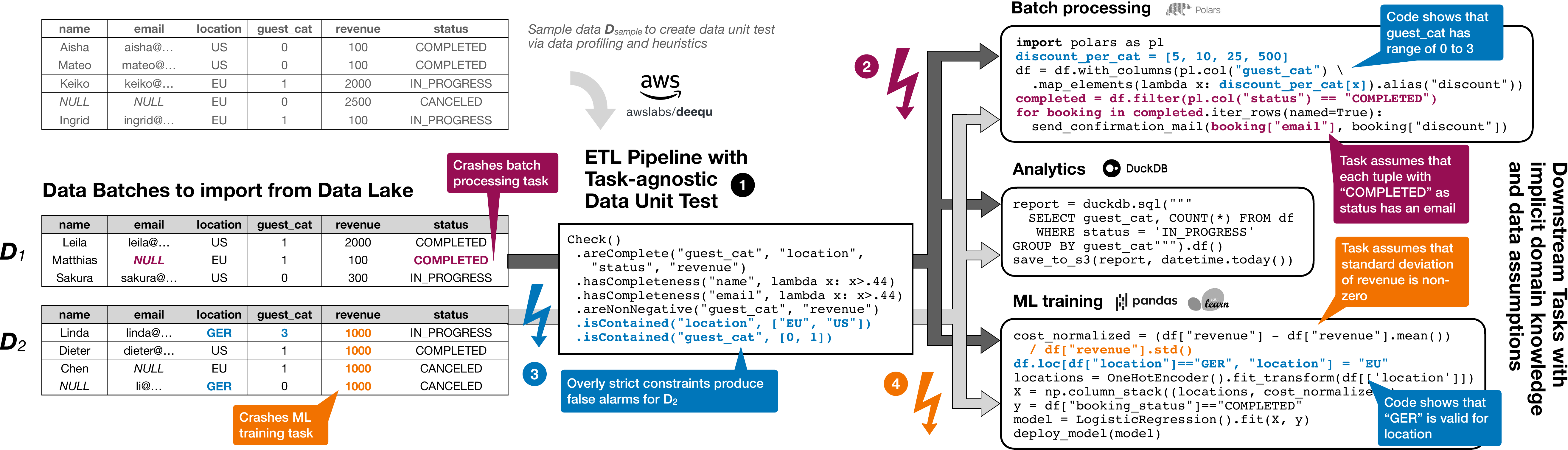}
\caption{Toy example to exemplify the need for task-aware data unit tests: \circled{1}~An ETL pipeline employs a task-agnostic data unit test (generated by AWS Deequ) to validate new batches of data before forwarding them to three downstream tasks. \circled{2}~A hidden dependency among different columns in the code of batch processing task causes a crash, and was missed by the Deequ test that only looked at data; \circled{3}~The overly strict data unit test flags data conditions to which downstream tasks are robust, and thereby causes false alarms; \circled{4}~A hidden assumption about the aggregate statistics of a column in the code of the ML task causes another crash. 
The example shows that a single data unit test derived from sample data alone is insufficient, since it fails to account for implicit data assumptions and domain knowledge in the code of the downstream tasks. Instead, a task-aware solution is required with a data unit test per downstream task, specialized to the task's access pattern and data assumptions.}
\label{fig:example}
\end{figure*}

\section{Problem Statement}
\label{sec:problem}

We discuss the shortcomings of task-agnostic data unit tests and introduce the problem in the focus of this paper with a running example in a fictitious scenario that mirrors the ETL and downstream-consumption patterns common in production data platforms. Note that we provide an executable version of this example in a Jupyter notebook at \textcolor{blue}{\url{https://github.com/deem-data/PrismaDV/blob/main/toy-example.ipynb}}.

\header{Running example} Imagine that a large travel corporation acquired a small startup which produced a successful booking app. As part of the integration, the central devops team from the corporation now needs to connect several downstream services of the startup with a large shared data lake from the corporation via ETL pipelines. These ETL pipelines regularly push new data into the downstream services (e.g., on a nightly basis). We visualise one such example pipeline in \Cref{fig:example}. This pipeline handles records,  which detail ongoing and completed bookings as well as their financial impact, with the following six columns \texttt{name}, \texttt{email}, \texttt{location}, \texttt{guest\_cat}, \texttt{revenue}, \texttt{status}. The ETL pipeline regularly feeds new batches of booking data into the following three downstream services developed by the startup:
\begin{itemize}[leftmargin=*]
\item \textit{Batch processing} -- a batch processing task which computes discounts for customers and sends them notification emails.
\item \textit{Analytics} -- a task which runs a SQL query to generate a daily report on active bookings and stores it in a distributed file system.
\item \textit{Machine learning model training} -- a task which uses the booking data to train and deploy a model to predict the probability of a booking completion.
\end{itemize}  

In the past, the devops team of the corporation has repeatedly had to handle data quality incidents where downstream tasks failed due to issues in the data, and the engineers had to spend their weekends fixing the data and rerunning the affected downstream tasks. \circled{1} To avoid such problems in this scenario, they decide to implement a data unit test for their ETL pipeline, which is evaluated on each new data batch to ingest, and is supposed to tell them whether it is safe to forward the newly arriving data. For that, they leverage the automated generation of data unit tests from Deequ (via ``constraint suggestion''~\cite{deequdoc}). The engineers take a sample $D_\text{sample}$ of the existing booking data, and provide it to Deequ. Deequ profiles the data sample and applies several heuristics to generate a data unit test in the form of a set of constraints on the completeness and value range of various columns. The devops team then deploys the generated test in their ETL pipeline.

\header{Reactive handling of data issues} At night, the data batch $D_1$ arrives in the ETL pipeline, which evaluates the data unit test on it. Since the test passes, the pipeline forwards to the data batch to the downstream tasks. \circled{2} However, the batch processing task crashes with an error, resulting in the devops team getting alerted. Their investigation uncovers that the code of the batch processing task contains the hidden assumption that each record with a ``COMPLETED'' value in the \texttt{status} must also have a valid value in the \texttt{email} column, which was not the case for the second tuple in $D_1$. This subtle condition has been missed by Deequ's constraint suggestion. The devops engineers now have to manually make sure that all customers receive their correct discount emails. Afterwards, they manually extend the data unit test to also account for the subtle data condition.

\begin{table*}[t!]
\centering
{\small
\begin{tabular}{p{1.8cm}|l|c|p{3.25cm}|p{6cm}}
\textbf{Module} & \textbf{API methods} & \textbf{LLM?} & \textbf{Output} & \textbf{Description}\\
\toprule

{\em Profiling \&} & \texttt{ProfileData} & \xmark & Data profile & Compute basic statistics about the input data \\
{\em Discovery} & \texttt{DiscoverColumnAccess} & \cmark & List of columns & Determine columns accessed by the downstream code\\
& \texttt{DiscoverJointColumnAccess} & \cmark & List of sets of columns & Determine columns jointly accessed by the downstream code \\
\midrule
{\em Assumption} & \texttt{ColumnDataflowAnalysis} & \cmark & Code locations & Find code lines operating on a column\\
{\em Inference} & \texttt{MultiColumnDataflowAnalysis} & \cmark &Code locations& Find code lines operating on a set of input columns\\
& \texttt{SummarizeAndLinkAssumptions} & \cmark & Data-code assumption graph & Summarize implicit data-code assumptions\\
\midrule
{\em Constraint Code} & \texttt{GenerateColumnConstraints} & \cmark & Executable constraint code & Generate constraints for a column\\
{\em  Generation}  & \texttt{GenerateMultiColumnConstraints} & \cmark &Executable constraint code & Generate constraints for a set of columns\\
\midrule
{\em Post-Processing}& \texttt{PreCheckConstraint} & \xmark & Flag indicating validity& discard buggy / invalid constraints\\
\bottomrule
\end{tabular}}
\caption{Modules and API methods of \system{}, implemented via external tools, custom code, and LLM invocations.}
\label{tab:api}
\end{table*}

During the next night, the data batch $D_2$ arrives in the pipeline. \circled{3} The data unit test rejects this batch, which leads to the quarantining of the data and again to alerts for the devops team. The engineers investigate the test results and find that the test flagged the unexpected value "GER" in the \texttt{location} column, as well as the value 3 for \texttt{guest\_cat}. After contacting the startup engineers, the devops team learns that this was a false alarm, the value "GER" is sometimes produced by legacy booking systems, and 3 is a rare but valid value for \texttt{guest\_cat}, which indicates a special guest category. The startup engineers confirm that both cases can be handled by their services, leading to the insight that the data unit test from Deequ was overly strict.
\circled{4} The engineers now make the ETL pipeline forward $D_2$, which unfortunately leads to an unexpected crash in the ML task. Investigating the code of the ML task uncovers that the data preparation code produces NaN values in the training data, which the ML model cannot handle. The devops engineers realize that this is due to the fact that \texttt{revenue} values are normalized by dividing through their standard deviation which is zero in this data batch. They again realize that this subtle data assumption was not covered in their data unit test. 

\header{The need for the automated generation of task-aware data unit tests} The examples show that a single central data unit test, derived from the data, is insufficient to adequately address potential data issues that can occur. Instead, an intricate understanding of the code and data assumptions of the downstream tasks are required. Ideally, a custom test for each downstream task, tailored to its specific data assumptions and access pattern is deployed. This would be to avoid both false alarms (which cause unnecessary work for devops engineers and on-call sessions on the weekend) and missed data issues (which may crash downstream services). However, creating specialized data unit tests is very tedious since popular datasets in large data lakes maybe consumed by hundreds of downstream services, often with hard-to-understand codebases (e.g., legacy code). Furthermore, both data and downstream services continuously change and evolve in large organizations, requiring a regular adjustment of the data unit tests.

\header{Research question} This leads us to the research question in the focus of this paper: {\em Can we automate the generation of data unit tests, such that they are tailored to downstream code?} Our goal is to change the development of data unit tests from its reactive nature (adjusting tests after production incidents) to a proactive nature~\cite{zeighami2025llm}, where comprehensive tests are generated upfront. At the same time, we aim to alleviate the need for domain experts to write custom data unit tests. An automated system should leverage task code in addition to sample data to design, specialize and improve tests, by uncovering and including the hidden domain knowledge and data assumptions in the code.

\header{Formal definition of task-aware data validation} We formalize the task-aware data validation problem introduced above. Consider a downstream task $T$---implemented as code artifact---that consumes a tabular dataset $D$ over time via regularly incoming data batches $\{D_1 , \dots, D_m\}$. We assume that $T$ executes correctly on a sample $D_\text{sample}$ of $D$; however, other batches may violate implicit assumptions embedded in the task logic, as illustrated in our running example. Whether $T$ succeeds or fails on a new data batch depends on whether these assumptions continue to hold.
We define the boolean \emph{task validity} of a data batch $D_i$ with respect to $T$ as $\mathsf{Valid}_T(D_i) \in \{0,1\} $, where $\mathsf{Valid}_T(D_i) = 1$ if executing $T$ on $D_i$ completes successfully and exhibits the intended behavior, and $\mathsf{Valid}_T(D_i) = 0$ if $T$ crashes, raises an exception, or silently produces an incorrect result. This outcome captures the ground-truth suitability of the data batch for the task.
The objective is to generate, for each downstream task $T$, a specialized data unit test in the form of a constraint set $C_T$ whose acceptance behavior aligns with the true task validity signal:
\[
    C_T(D_i) \Leftrightarrow \mathsf{Valid}_T(D_i),
\]
for both observed and, critically, \emph{unobserved} dataset batches.
Given the downstream task code $T$ and an observed data sample $D_\text{sample}$ of $D$ on which $T$ runs successfully, the goal is to infer the implicit data assumptions that $T$ relies on to operate correctly and synthesize from them a constraint set $C_T$ that approximates the validity function $\mathsf{Valid}_T(\cdot)$ on new data batches.


\section{PrismaDV}
\label{sec:approach}

\begin{figure*}[t!]
\centering
\includegraphics[width=\textwidth]{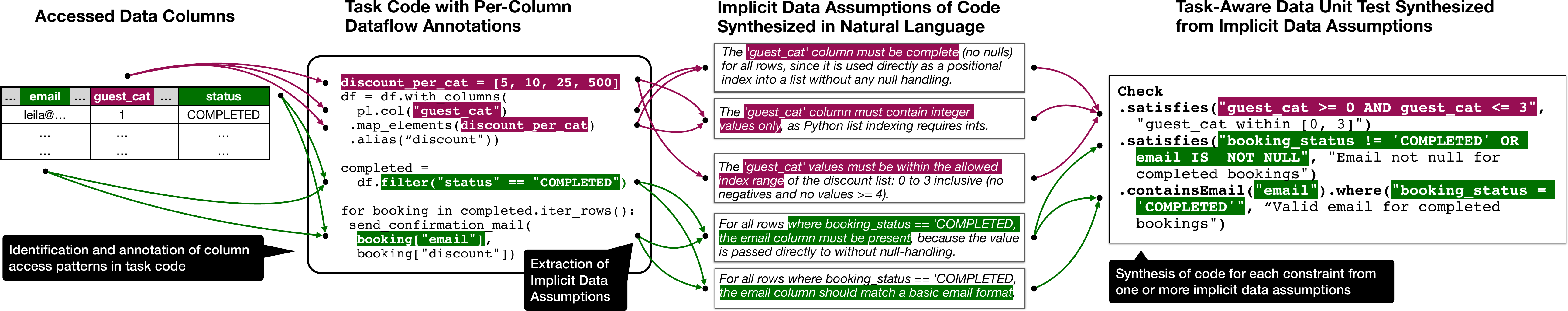}
\caption{During data unit test generation for a downstream task, \system{} builds a bipartite ``data-code assumption'' graph, which connects accessed input columns to the implicit data assumptions in the task code about them (synthesized in natural language). For that, our system annotates the code lines that operate on an input column (or data derived from it). The code generation module, which synthesizes the task-aware data unit test, leverages the assumption graph as input. }
\label{fig:approach}
\end{figure*}

In \system{}, we leverage Large Language Models (LLMs) for task code summarization, assumption inference, and constraint code generation, since these models have recently shown strong capabilities in code generation~\cite{fathollahzadeh2025demonstrating, bassamzadeh2024comparative, shankar2024docetl}, data preprocessing~\cite{FanFTCLD25, flokas2025towards, 10.1145/3722212.3725124,10.1145/3627673.3679216, 10.1145/3448016.3452750,li2025jupiter}, and program understanding~\cite{nam2024using, 10.14778/3685800.3685835,perez-etal-2025-llm}. Note that our modular API isolates LLM interactions within specific methods.

\subsection{System Modules} 

We decompose task-aware data unit test generation into three modules, each exposing a well-defined API contract: (i)~{\em Profiling and Discovery}, (ii)~{\em Assumption Inference}, (iii)~{\em Constraint Code Generation}, and (iv)~{\em Post-processing}. Each module exposes a set of API methods for the data unit test generation workflow detailed in Table~\ref{tab:api}. Formally, \system{} takes as input a data sample $D_\text{sample}$ with columns $A = [A_1, \dots, A_n]$ and the source code of the downstream task $T$.

\header{Profiling and discovery} The profiling and discovery module collects descriptive statistics from sample data and analyzes the code of downstream tasks to identify the columns, and combinations of columns, accessed by downstream tasks. This establishes the foundation for connecting data characteristics with task semantics.

Given $D_\text{sample}$ and $T$, the method \texttt{ProfileData}$(D_\text{sample}) \rightarrow S$ computes descriptive statistics $S$, which include types, completeness, approximate number of distinct values, histograms for low-cardinality columns, and the mean for numeric columns. The methods \texttt{DiscoverColumnAccess}$(T, \mathrm{A}, S) \rightarrow A_\text{accessed}$ and \texttt{DiscoverJointColumnAccess}$(T, A_{\text{accessed}}, S) \rightarrow A_\text{accessed\_jointly}$ detect the subset of columns accessed by $T$ and column groups jointly referenced in the code, respectively. The resulting metadata provides the basis for the subsequent assumption inference stage.

\header{Assumption inference and data-code graph construction} The assumption inference module forms the conceptual core of \system{}. It bridges the gap between data and code by analyzing the task code to derive explicit representations of the implicit data assumptions encoded within. This module transforms the task code into a structured, interpretable intermediate representation, producing natural language descriptions of these hidden assumptions. The assumption inference module constructs a bipartite \emph{data–code assumption graph} $G = ((A_\text{relevant}, H), E)$, where $A_\text{relevant} = A_\text{accessed} \cup A_\text{accessed\_jointly}$ denotes accessed columns, $H$ the set of inferred data assumptions, and $E$ the labeled edges linking them, annotated with code locations. For each column $A_i \in \mathrm{A}_{\text{accessed}}$, \texttt{ColumnDataflowAnalysis}$(T, A_i) \rightarrow L_i$ locates the statements in $T$ operating on $A_i$ or its derivatives. The code locations $L_i$ are then used to create an annotated code variant $T^\prime$ of the code $T$. This annotated code is then summarized through \texttt{summarizeAndLinkAssumptions}$(T^\prime, A_i, S)$, yielding the set of inferred natural language assumptions $H_i$ connected to code spans in $L_i$. For multi-column cases, \texttt{MultiColumnDataflowAnalysis}$(T, A_j)$ identifies joint access of column sets $A_j \in \mathrm{A}_{\text{accessed\_jointly}}$ in the code. The resulting graph $G$ serves as the \emph{intermediate representation} passed to the constraint synthesis stage. We refer to \Cref{fig:approach} for a visualization of this process on a downstream task from our running example.

\header{Constraint code generation} The constraint code generation module synthesizes executable validation logic from the data–code assumption graph $G$, translating the inferred data assumptions linked to code and columns directly into the syntax of a target data validation framework (e.g., Deequ or Great Expectations). The methods \texttt{GenerateColumnConstraints}$(A_i, G)$ and \texttt{GenerateMultiColumnConstraints}$(A_j, G)$ translate the data-code assumption graph $G$ into executable constraints for the accessed columns $A_\text{relevant}$ expressed in the syntax of a target data validation framework. Multiple constraints may arise from a single assumption, and conversely, one constraint may aggregate several related assumptions.

\header{Post-processing} The last module ensures syntactic validity of the generated constraints. Each candidate constraint $c$ is first validated via \texttt{PreCheckConstraint}$(c, D_\text{sample}) \rightarrow \{0, 1\}$, which evaluates parseability and consistency. Secondly, constraints that do not hold on $D_{\text{sample}}$ are discarded.

\header{Implementation} We implement the proposed modules of \system{} in DSPy~\cite{khattabdspy} with support for response caching and asynchronous execution to run dataflow analysis, assumption generation, and \texttt{pydeequ} code generation in parallel across accessed columns.


\section{Optimization via Selective Informative Feedback for Task Adaptation (SIFTA)}
\label{sec:approach_optimization}

In real-world deployments, data validation runs as part of data pipelines that continuously ingest new data batches. Over time, teams accumulate execution outcomes that indicate whether a task run succeeded or failed on specific batches. These observations are scarce, but they provide feedback for improving validation quality on future data. Moreover, multiple tasks often consume the same input dataset. They can share latent data assumptions or business logic, such as preprocessing steps, joins, or feature engineering. This creates transfer opportunities: outcomes from existing tasks and batches can help improve validation for new batches and tasks.

Taken together, these properties motivate an optimization approach for task-aware data unit tests generation. Since \system{} is a compound AI system with LLM-based modules, there are several optimization choices such as updating LLM parameters, e.g., via reinforcement learning or fine-tuning or adjusting prompts of the LLM-based modules. In production settings, however, updating LLM parameters for each new task is often infeasible, as it incurs significant training cost, fine-tuning parameter storage, and deployment and versioning overhead~\cite{10.1145/3769776}. We therefore focus on prompt optimization for \system{}'s LLM-based modules, as it requires no model training, fits within existing inference pipelines, and can be driven by the scarce execution outcomes collected in production.

\subsection{Optimization Setting}
\label{sec:optimization_problem}

We extend the formal problem statement introduced in \Cref{sec:problem}. Over time, downstream tasks are executed on new data batches, yielding task--batch pairs with an observed binary execution outcome indicating whether the task completed correctly. Although these observations are scarce and expensive to obtain, they provide valuable feedback for adapting the system to future data batches and downstream tasks. Our goal is to tune the prompts $\Pi$ of the LLM-based modules used by \system{} for a fixed dataset, using observed execution outcomes together with the corresponding data batches and task code. We consider \emph{three} within-dataset generalization settings: (i) new batches for known tasks, (ii) new tasks on known batches, and (iii) new tasks on new batches.

Formally, for dataset $D$ with $m$ data batches $\{D_i\}_{i=1}^m$, let $\{T_\ell\}_{\ell=1}^{r}$ be the $r$ downstream tasks that consume the data over time. Let $T_{\mathrm{obs}}$ denote known observed tasks, and $T_{\mathrm{new}}$ denote new tasks for the same dataset. Analogously, let $D_{\mathrm{obs}}$ denote observed data batches and $D_{\mathrm{new}}$ unseen new data batches. The observation set is $\mathcal{O} \subseteq T_{\mathrm{obs}} \times D_{\mathrm{obs}}$. \system{}'s LLM-based modules depend on a backbone LLM $L$ and the set of prompts $\Pi$. Concretely, we optimize $\Pi$ while keeping $L$ fixed; we omit $L$ from the notation for readability.
The optimization target $\mathcal{Q}$ can be defined in different ways, covering (i) new data batches $D_{\mathrm{new}}$ for observed tasks $T_{\mathrm{obs}}$, (ii) new tasks $T_{\mathrm{new}}$ on observed batches $D_{\mathrm{obs}}$, or even (iii) new tasks $T_{\mathrm{new}}$ on new data batches $D_{\mathrm{new}}$. For a given set of prompts $\Pi$, \system{} generates for each task $T_\ell$ a task-specific data unit test $C_\ell^{(\Pi)}$, whose evaluation on data batch $D_i$ yields a binary prediction $\widehat{v}_{\ell,i}^{(\Pi)} = C_\ell^{(\Pi)}(D_i)$. We denote the actual binary execution outcome for a task--batch pair as $v_{\ell,i} = \mathsf{Valid}_{T_\ell}\!\left(D_i\right)$; note that these ground-truth outcomes are only observed for deployed data unit tests.

\header{Optimization objective} Our goal is to choose a set of prompts $\Pi$ that maximize validation quality on the target scenarios $\mathcal{Q}$.
Formally, the objective is:
\[
    \Pi^{*}
    ~=~
    \arg\max_{\Pi}
    \;
    \mathbb{E}
    \Bigl[
        \mu\bigl(
        \{(\widehat{v}_{\ell,i}^{(\Pi)}, v_{\ell,i}) : (T_\ell, D_i) \in \mathcal{Q}\}
        \bigr)
        \Bigr],
\]
where $\mu(\cdot)$ is a validation-quality metric such as F1 score.

\header{Challenges} The optimization problem is challenging due the following reasons:

\headerl{Scarce and delayed supervision} The primary feedback available in practice is a binary execution outcome for a task run on a data batch. Obtaining more specific supervision signals (e.g., which column caused a failure or which assumption was violated) is difficult and expensive, and may require substantial debugging. This issue is exacerbated in ML pipelines, where the manifestation of data issues can be delayed (e.g., gradual performance degradation), making it hard to collect fine-grained error feedback at scale.

\headerl{Multiple intermediate trajectories with localized errors} Task-aware validation generates rich intermediate artifacts (e.g., column access patterns, assumption summaries, constraint candidates, and code), which can be long and heterogeneous. At the same time, data issues in a batch are typically localized to a small subset of columns or interactions. This mismatch makes it hard to directly apply existing prompt-optimization methods (e.g., MIPROv2~\cite{opsahl-ong-etal-2024-optimizing}, GEPA~\cite{agrawal2025gepa}), which are commonly evaluated in settings with smaller trajectories and more dense feedback.

\subsection{``Failure Precision'' as Informative Signal}

\header{Informative learning signals}
Task-aware data validation generates multiple signals from the execution of constraints and downstream tasks. However, the available supervision is limited to binary execution outcomes, and not all observed signals are equally informative for assessing the quality of a constraint. We analyze under which conditions the outcome of a constraint evaluation provides reliable information about whether the constraint captures task-relevant data errors.

We consider the behavior of an individual constraint. Let $c_{\ell,k}$ denote a constraint defined on column $A_j$, evaluated on a data batch $D_i$. For each evaluation, we observe the binary constraint outcome $\widehat{v}_{\ell,i,k} \in {0,1}$ and the corresponding task execution outcome $v_{\ell,i} \in {0,1}$. Whether $A_j$ contains a task-relevant data error in batch $D_i$ is not directly observable and is treated as a latent variable.

Based on the observable outcomes, four cases can occur:
(1)~$\widehat{v}_{\ell,i,k}=1$ and $v_{\ell,i}=1$: the constraint passes and the task succeeds. This outcome is inconclusive, as the absence of observed failures does not imply that the constraint would detect relevant errors in other batches;
(2)~$\widehat{v}_{\ell,i,k}=1$ and $v_{\ell,i}=0$: the constraint passes while the task fails. This case is also ambiguous, since the task failure may be caused by errors in other columns or by interactions that the constraint does not capture;
(3)~$\widehat{v}_{\ell,i,k}=0$ and $v_{\ell,i}=1$: the constraint fails while the task succeeds. This outcome corresponds to a clear false alarm and therefore provides a reliably informative negative signal;
and (4)~$\widehat{v}_{\ell,i,k}=0$ and $v_{\ell,i}=0$: both the constraint and the task fail. This outcome is potentially informative, but remains ambiguous because the task failure may or may not be attributable to an error in column $A_j$.

The key asymmetry arises when a task fails ($v_{\ell,i}=0$): there are two latent possibilities, namely that the failure is caused by a task-relevant error in column $A_j$, or that it originates from other columns or interactions. Since this distinction is unobserved, multiple latent error configurations collapse into the same observable outcome. Consequently, constraint passes ($\widehat{v}_{\ell,i,k}=1$) are inherently uninformative under this supervision regime, as they are compatible with both the absence of errors and undetected errors. In contrast, constraint failures ($\widehat{v}_{\ell,i,k}=0$) are the only outcomes that can yield informative learning signals: failures with task success indicate definitive false alarms, while failures with task failure correspond to plausible detections. This observation motivates focusing optimization exclusively on constraint failures and quantifying how often a constraint failure coincides with a task failure.

\header{Column-level failure precision}
Our optimization operates on task--column units. For a column $A_j$, let $C_{\ell,A_j}$ be the set of constraints on $A_j$ and define the column-level prediction $\widehat{w}_{\ell,i,j}$ on a batch $D_i$ as $\widehat{w}_{\ell,i,j} = \bigwedge_{C_{\ell,k} \in C_{\ell,A_j}} \widehat{v}_{\ell,i,k}$. The \emph{column-level failure precision} is an empirical estimate of $\Pr[v_{\ell,i}=0 \mid \widehat{w}_{\ell,i,j}=0]$ over observed batches $D_{\mathrm{obs}}$:
\[
    \mathrm{CFPr}(C_{\ell}, A_j, D_{\mathrm{obs}}) =
    \frac{\sum_{D_i \in D_{\mathrm{obs}}} \mathbf{1}[\widehat{w}_{\ell,i,j} = 0 \wedge v_{\ell,i} = 0]}{\sum_{D_i \in D_{\mathrm{obs}}} \mathbf{1}[\widehat{w}_{\ell,i,j} = 0]},
\]
and we only consider columns with non-zero denominator as informative training units.

\header{Constraint-level failure precision}
For diagnosis and backtracing within an informative column, we also compute constraint-level failure precision for individual failing constraints:
\[
    \mathrm{FPr}(C_{\ell,k}, D_{\mathrm{obs}}) =
    \frac{
        \sum_{D_i \in D_{\mathrm{obs}}} \mathbf{1}\!\left[\widehat{v}_{\ell,i,k} = 0 \wedge v_{\ell,i} = 0\right]
    }{
        \sum_{D_i \in D_{\mathrm{obs}}} \mathbf{1}\!\left[\widehat{v}_{\ell,i,k} = 0\right]
    }.
\]
Constraints that never fail (zero denominator) are assigned $\mathrm{FPr}=0$, since they provide no actionable information.

\subsection{Optimization Procedure}

Based on the analysis above, we introduce \emph{Selective Informative Feedback for Task Adaptation} (SIFTA), a lightweight prompt-optimization procedure that leverages informative failure signals captured by failure precision. SIFTA concentrates optimization on task--column units with constraint failures and backtraces failing constraints to the assumptions and code locations that produced them, providing targeted feedback for prompt updates.

\header{Overview} SIFTA iteratively updates PrismaDV's constraint-generation related prompts $\pi$, i.e., the prompts used by the LLM-based API methods \texttt{ColumnDataflowAnalysis}, \texttt{SummariseAndLinkAssumptions}, and \texttt{GenerateColumnConstraints} (Table~\ref{tab:api}), using scarce task outcomes across multiple tasks on the same dataset. At the beginning of each round, SIFTA (i) condenses the training observations using the current $\pi$, (ii) samples $n_{\mathrm{eval}}$ task--column units for evaluation, and scores the current $\Pi$ on this fixed evaluation sample, and (iii) allocates the remaining evaluation budget $b_{\mathrm{eval}}$ across rounds. Within a round, it repeatedly resamples $n_{\mathrm{train}}$ training units to generate constraints, compute failure-precision scores, and construct backtraces; a candidate $\Pi'$ is only scored on the round's evaluation sample if its mean training $\mathrm{CFPr}$ on the sampled training units does not decrease. \Cref{alg:prompt-optimization} shows the procedure.

\header{Training set condensation} To reduce the search space, we first only generate candidate constraints for the actual accessed columns of a task $T_\ell$. At the beginning of each optimization round, we condense the training set $\mathcal{O}_{\mathrm{train}}$ using the current prompts $\Pi$ by selecting task--column units $(T_\ell, A_j)$ for which at least one constraint fails on an observed training batch. This concentrates the optimization budget on columns that surface failures. We do not condense $\mathcal{O}_{\mathrm{eval}}$; instead, we uniformly sample task--column units from all task--column combinations in $\mathcal{O}_{\mathrm{eval}}$ and keep this eval sample fixed within the round.

\header{Selection of training targets via failure precision} We use mean column-level failure precision $\mathrm{CFPr}$ as the primary optimization objective. For each sampled training unit, we compute constraint-level failure precision $\mathrm{FPr}$ for failing constraints and rank them within each column. For feedback, we select the $n_{\mathrm{fb}}$ constraints with the lowest $\mathrm{FPr}$ per column, emphasizing negative signals that are directly actionable for prompt updates.

\header{Backtracing feedback context} For each selected low-$\mathrm{FPr}$ constraint, we backtrace to the linked assumptions and code locations via the data--code assumption graph (\Cref{fig:approach}), and provide these traces (together with $\mathrm{CFPr}$ and $\mathrm{FPr}$ scores) as feedback context to the prompt proposer. The proposer returns a candidate prompt set $\Pi'$, which we score on the evaluation sample only if its mean training $\mathrm{CFPr}$ on the sampled units does not decrease.

\header{Implementation} We implement SIFTA using DSPy's prompt-optimization API. In our pipeline, the prompts for dataflow analysis, assumption inference, and constraint code generation are tightly coupled through shared intermediate artifacts. As a result, updating only one module prompt can create contextual mismatches across modules. SIFTA therefore uses a global prompt proposer that can jointly update one or multiple module prompts within a single proposal. Each proposal is conditioned on a shared instruction prompt that provides task-aware data validation context and motivates failure precision as the proxy optimization target.

\begin{algorithm}[t]
    \caption{Prompt optimization for \system{} with column-level failure-precision backtracking via SIFTA.}
    \label{alg:prompt-optimization}
    \begin{algorithmic}[1]
        \Statex \textbf{Require:} training observations $\mathcal{O}_{\mathrm{train}}$, eval observations $\mathcal{O}_{\mathrm{eval}}$, initial constraint-generation prompts $\Pi_0$, rounds $n_{\mathrm{round}}$, eval budget $b_{\mathrm{eval}}$, train sample size $n_{\mathrm{train}}$, feedback constraints per column $n_{\mathrm{fb}}$, eval sample size $n_{\mathrm{eval}}$, assumption graph $G$
        \State $\Pi \gets \Pi_0$
        \State $b_{\mathrm{remain}} \gets b_{\mathrm{eval}}$
        \For{$t \gets 1 \dots n_{\mathrm{round}}$}
            \State $\mathit{trainCond} \gets \textsc{Condense}(\mathcal{O}_{\mathrm{train}}, \Pi)$
            \State $\mathit{evalSample} \gets \textsc{SampleColumns}(\mathcal{O}_{\mathrm{eval}}, n_{\mathrm{eval}})$
            \State $\mathit{evalScore} \gets \textsc{MeanCFPr}(\Pi, \mathit{evalSample})$
            \State $b_t \gets \left\lfloor b_{\mathrm{remain}} / (n_{\mathrm{round}} - t + 1) \right\rfloor$
            \State $\mathit{candidates} \gets \{(\Pi, \mathit{evalScore})\}$
            \While{$b_t > 0$}
                \State $\mathit{trainSample} \gets \textsc{SampleColumns}(\mathit{trainCond}, n_{\mathrm{train}})$
                \State $\mathit{trainScore} \gets \textsc{MeanCFPr}(\Pi, \mathit{trainSample})$
                \State $\mathit{colCFPr} \gets \textsc{ComputeCFPr}(\Pi, \mathit{trainSample})$
                \State $\mathit{constraintFPr} \gets \textsc{ComputeFPr}(\Pi, \mathit{trainSample})$
                \State $\mathit{lowFPr} \gets \textsc{SelectBottomKPerColumn}(\mathit{constraintFPr}, n_{\mathrm{fb}})$
                \State $\mathit{traces} \gets \textsc{Backtrace}(\mathit{lowFPr}, G)$
                \State $\Pi' \gets \textsc{Propose}(\Pi, \mathit{colCFPr}, \mathit{lowFPr}, \mathit{traces})$
                \If{$\textsc{MeanCFPr}(\Pi', \mathit{trainSample}) \ge \mathit{trainScore}$}
                    \State $\mathit{evalScore}' \gets \textsc{MeanCFPr}(\Pi', \mathit{evalSample})$
                    \State $\mathit{candidates} \gets \mathit{candidates} \cup \{(\Pi', \mathit{evalScore}')\}$
                    \State $b_t \gets b_t - 1$; $b_{\mathrm{remain}} \gets b_{\mathrm{remain}} - 1$
                \EndIf
            \EndWhile
            \State $(\Pi, \mathit{evalScore}) \gets \textsc{BestByEval}(\mathit{candidates})$
        \EndFor
        \State \Return $\Pi$
    \end{algorithmic}
\end{algorithm}

\section{Benchmarking Task-Aware Data Validation}
\label{sec:benchmarks}

We introduce \icdbench{} and \eidbench{}, two carefully designed benchmarks to evaluate task-aware data unit tests generation. These benchmarks provide a standardized way to compare future frameworks as well as various baselines such as LLMs, outlier detection methods and agentic systems~\cite{DBLP:journals/corr/abs-2402-01108,summers2025please} in terms of their ability to automatically generate effective, task-aware data unit tests. Our benchmarks integrate publicly available datasets with LLM-generated tasks covering diverse domains and applications. We make both benchmarks available under an open license at \textcolor{blue}{\url{https://github.com/deem-data/PrismaDV/blob/main/benchmarks}} and plan to maintain and update them with new use cases and baselines.

\subsection{\icdbench{} -- Individual Constraint Discovery from Data-Code Pairs}
\label{sec:benchmarks-icdb}

The building block of a task-aware data unit test generation system is the ability to discover constraints about the data which are implicitly defined in code. To evaluate this ability (independent of the other end-to-end building blocks of the system), we design \icdbench{}, a novel benchmark for constraint discovery from data-code pairs. This benchmark includes 63 cases, each of which consists of a tabular data sample, a piece of code written to process the data, a natural language description of the hidden assumption in the code, and the corresponding ground truth constraint in PyDeequ syntax. Furthermore, each case features two held-out pieces of data: a positive example of data to pass (on which the constraint holds), and a negative example of data to reject, where the constraint is not satisfied. 
\icdbench{} includes hand-designed cases as well as a large number of data-code pairs obtained from public GitHub repositories. We explicitly design it to include a diverse range of hidden assumptions in code, ranging from simple cases like explicit asserts, to tough cases like column dependencies expressed in control flow or knowledge about semantics in ML libraries (e.g., scikit-learn operations). Moreover, the benchmark covers a wide range of domains, including payment processing, cricket sports rules, and in-game auctions in video games.

\subsection{\eidbench{} -- End-to-End Error Impact Detection}
\label{sec:benchmarks-eidb}

Following the task-aware data validation setting introduced in \Cref{sec:problem} (a downstream task $T$ consuming a dataset over data batches $\{D_1,\dots,D_m\}$), we introduce \eidbench{}, an end-to-end benchmark for evaluating full pipelines. Again, a batch is ``safe'' for $T$ only if executing $T$ completes and exhibits the intended behavior; otherwise it is labeled ``erroneous''.

\header{Benchmark design} Each \eidbench{} dataset provides: (i) an initial \emph{data sample} $D_\text{sample}$ used to author and validate task code, (ii) a set of twenty-five \emph{evaluation batches}, each obtained by injecting synthetic errors of a certain type into $D_\text{sample}$, and (iii) a suite of downstream tasks in Python written to consume the dataset. The code of each task embeds \emph{data assumptions} as executable assertion blocks, which we use as ground truth to label whether an evaluation batch is safe or erroneous for that specific task.

\header{Datasets} We include five datasets from diverse domains and sources, with varying fractions of numerical, categorical, and textual attributes.

\header{Downstream tasks} A core challenge in building an end-to-end benchmark is obtaining diverse,
executable downstream tasks for the same tabular dataset: such code is common in industry but rarely shared with the academic community, while public notebooks, like the ones from Kaggle, often follow repetitive EDA/model-training templates. We therefore synthesize downstream tasks and their ground-truth assumptions using an LLM-assisted,
human-in-the-loop pipeline.

\headerl{Task generation pipeline} For each dataset, we treat the sample data $D_\text{sample}$ as the development data
and generate an initial pool of candidate tasks via four stages:
\begin{enumerate}[leftmargin=*,noitemsep]
    \item \textit{Table summarization}: we profile $D_\text{sample}$ and produce a compact
    summary of its schema and value distributions (types, missingness, ranges,
    frequent categories, and example rows).
    \item \textit{Task proposal}: conditioned on the summary, the LLM proposes a
    concrete task description that mimics applied business logic over the
    table.
    \item \textit{Assumption generation}: the LLM enumerates implicit data
    assumptions that a developer would rely on for the task to behave
    correctly. Each assumption is phrased as a predicate whose violation
    would lead to a crash or an abnormal (possibly silent) behavior.
    \item \textit{Code generation}: the LLM implements the task as a single
    Python script. We prompt it to (i) rely on the assumptions in the task
    logic and (ii) embed each assumption as an executable assertion block (see
    below).
\end{enumerate}
We generate 30 candidate tasks per dataset and retain the executable tasks after the verification procedure described below.

\header{Assertion blocks and leakage control} Each assertion block is delimited by sentinel comments (e.g., \texttt{\# ASSERTION\_START} / \texttt{\# ASSERTION\_END}) to enable programmatic removal and reinsertion. We use the blocks in two ways: (i) for \emph{inference}, we remove all assertion blocks and provide the resulting code to the system under test, preventing trivial extraction of assumptions from explicit asserts; (ii) for \emph{labeling}, we execute the script with assertion blocks enabled on evaluation batches. A data batch is labeled \emph{erroneous} for a task if its execution crashes or any assertion fails; otherwise it is labeled \emph{safe}. To prevent leakage, we ensure that removing assertion blocks leaves the task executable and that no program state used by the core task is defined or modified inside assertion blocks. In addition, we manually review each task's code to remove bugs and eliminate assumption leakage via comments or task logic.

\header{Task selection and verification} Each generated task goes through two verification stages: (i) an automated
repair-and-test loop that runs the task on $D_\text{sample}$ in three modes: (a) with all assertion blocks enabled, (b) with all assertion blocks removed, and (c) with exactly one assertion block enabled at a time (removing all others), to ensure that each assumption check is executable and independent. This guards against implementation artifacts where the core task code accidentally relies on variables, imports, or intermediate results created or redefined inside assertion blocks, which would break when assertions are removed. Upon failure, we prompt the same LLM to edit the code up to five rounds, discarding tasks that remain non-executable; (ii) a manual audit to remove remaining bugs, confirm expected behavior on $D_\text{sample}$, and check for any assumption leakage. Following this procedure, we retain 60 final tasks (roughly 12 per dataset). We release the initially generated task versions, edit histories, and final tasks with the benchmark.

\header{Error injection}
To mimic real-world data issues, we extend the tabular error injection framework Jenga~\cite{schelter2021jenga} with 19 operator types covering structural, integrity, numerical, textual, and format corruptions. For each dataset, we instantiate these operators into 25 error configurations, each producing one corrupted data batch from the clean sample $D_\text{sample}$. Each configuration corrupts only a small subset of columns and rows, reflecting how production issues are typically localized. We design task-targeted corruptions by inspecting scripts' assumption blocks so that a given corruption may break only a subset of scripts while leaving others unaffected. This setting challenges task-aware data validation methods, which must catch harmful corruptions for affected tasks while avoiding false alarms for robust tasks. All error configurations are released with the benchmark in our repository.

\header{Evaluation} Each pair of (task, evaluation batch) must be classified as \emph{pass} (safe) or \emph{reject} (erroneous). We compare these predictions to the ground-truth labels produced by executing tasks with assertion blocks enabled, and report precision, recall, and F1 score for detecting erroneous batches.


\section{Related Work}
\label{sec:related_work}

\header{Data validation}
Existing data validation systems vary in how rules are specified and inferred. Great Expectations offers a flexible assertion grammar but relies on manually defined expectation suites, limiting automation. Deequ~\cite{schelter2018automating} and TFDV~\cite{shankar2023automatic} infer statistical constraints via data profiling, while Auto-Test~\cite{10.1145/3725396} and Auto-Validate~\cite{song2021auto} learn semantic constraints from large table corpora, with Auto-Validate focusing on string columns. These approaches are largely task-agnostic and depend primarily on observed data. DataPrism~\cite{galhotra2022dataprism} incorporates downstream systems, using causal reasoning to identify data-profile violations that trigger failures. In contrast, \system{} generates task-aware validation rules across heterogeneous columns by jointly reasoning over data and code, capturing both explicit violations and latent data issues that induce abnormal program behavior.

\header{Code understanding with LLMs} \system{}'s performance in extracting data assumptions depends on the code understanding capabilities of LLMs~\cite{nam2024using, 10.1145/3597503.3639187, repoqa, nguyencodemmlu, jimenez2023swe}. Recent studies have demonstrated that LLMs can reason about code execution behavior and program semantics~\cite{chen2024reasoning, liu2025tool, liu2024codemind}. LLMDFA~\cite{wang2024llmdfa} further shows that LLMs can serve as effective tools for performing data-flow analysis over source code, and RepoAudit~\cite{guo2025repoaudit} leverages such reasoning for repository-level auditing. Multiple benchmarks evaluate LLM code understanding and reasoning abilities~\cite{jelodar2025large, dehghan2024assessing, lu1codexglue, gu2024cruxeval}. However, none of these benchmarks explicitly target implicit data assumptions embedded in code, which our \icdbench{} and \eidbench{} capture (\Cref{sec:benchmarks}).

\header{Prompt optimization for compound AI systems} LLM performance can vary substantially with prompt quality. Methods such as Chain-of-Thought~\cite{wei2022chain} and Plan-and-Solve prompting~\cite{wang2023plan} have been shown to improve reasoning performance. Beyond manual prompt design, prompt optimization methods~\cite{zhou2022large, yang2023large, fernando2023promptbreeder, gurajada2025effectiveness} aim to automatically improve prompts using feedback from previous invocations. Inspired by PyTorch’s abstraction philosophy~\cite{paszke2019pytorch}, DSPy~\cite{khattabdspy} offers a declarative framework for defining and optimizing prompt modules using textual feedback. Building on DSPy, MIPROv2~\cite{opsahl-ong-etal-2024-optimizing} selects high-performing instructions and demonstrations via Bayesian optimization. GEPA~\cite{agrawal2025gepa} evolves prompts with Pareto-based selection and can outperform reinforcement learning methods such as GRPO. TextGrad~\cite{yuksekgonul2025optimizing} treats prompts and intermediate outputs as optimizable variables and updates them via backpropagated natural-language feedback. EvoPrompt~\cite{guoconnecting} applies evolutionary search to prompt optimization, while AlphaEvolve~\cite{novikov2025alphaevolve} extends this idea to code through execution-based evaluation.
\section{Experimental Evaluation}
\label{sec:experiments}

We experimentally evaluate \system{} in the following. We start by assessing its ability to discover individual constraints from data-code pairs (\Cref{sec:exp-icdb}) and the ability of its tests to accommodate for the end-to-end impact of errors (\Cref{sec:exp-end-to-end}), based on our benchmarks from \Cref{sec:benchmarks}. Next, we evaluate the ability of our proposed SIFTA approach to optimize the prompts of our system in \Cref{sec:exp-optimization}. Finally, we conduct an ablation study in \Cref{sec:exp-ablation} to quantify the impact of our individual system modules.


\subsection{Constraint Discovery from Data-Code Pairs}
\label{sec:exp-icdb}

\begin{table*}[h!]
\centering
{\small
\setlength\extrarowheight{-1pt}
\begin{tabular}{lll|c|cccc|cc}
 & Task- & Generates & Avg. num.& \multicolumn{2}{c}{\textbf{Data to Pass}} & \multicolumn{2}{c}{\textbf{Data to Reject}} & False alarm & \\
\textbf{Method} & aware? & test?& const.& Passed$\uparrow$ & False alarm$\downarrow$ & Rejected$\uparrow$ & Missed$\downarrow$ & or missed$\downarrow$  & \textbf{F1 Score}$\uparrow$ \\ \toprule
\texttt{one-class-svm} & - & - & - & \phantom{0}0 & 63 & 61 & \phantom{0}2 & 65 & \phantom{0}0.0\% \\  
\texttt{isolation-forest} & - & - & - & 14 & 49 & 51 & 12 & 61 & 31.5\% \\  
\texttt{stats-novelty} & - & - & - & \phantom{0}0 & 63 & 63 & \phantom{0}0 & 63 & \phantom{0}0.0\% \\  
\midrule
\texttt{deequ} &  -& \checkmark & 1.9 & 44 & 19 & 27 & 36 & 55 & 61.5\%\\ 
\texttt{tensorflow-dv} & -  & \checkmark & - & 32 & 31 & 32 & 31 & 62 & 50.8\% \\
\midrule
\texttt{zero-shot [gemini-2.5-flash]} &  \checkmark & \checkmark & 2.4 & 22 & 41 & 59 & \phantom{0}4 & 45 & 49.4\%\\
\texttt{zero-shot [gemini-2.5-pro]} &  \checkmark & \checkmark & 2.8 & 19 & 44 & 60 & \phantom{0}3 & 47 & 44.7\%\\
\texttt{zero-shot [gpt-4.1]} & \checkmark  & \checkmark & 3.1 & 19 & 44 & 56 & \phantom{0}7 & 51 & 42.7\% \\
\texttt{zero-shot [gpt-5-mini]} & \checkmark  & \checkmark & 3.5 & 30 & 33 & 57 & \phantom{0}6 & 39 & 60.6\% \\
\texttt{zero-shot [gpt-5]} & \checkmark  & \checkmark & 3.1 & 30 & 33 & 57 & \phantom{0}6 & 39 & 60.6\% \\
\midrule
\texttt{few-shot [gemini-2.5-flash]} & \checkmark  & \checkmark & 2.0 & 27 & 36 & 51 & 12 & 48 & 52.9\% \\
\texttt{few-shot [gemini-2.5-pro]} & \checkmark  & \checkmark & 1.9 & 34 & 29 & 54 & \phantom{0}9 & 38 & 64.2\% \\
\texttt{few-shot [gpt-4.1]} & \checkmark  & \checkmark & 2.1 & 29 & 34 & 52 & 11 & 45 & 56.3\% \\
\texttt{few-shot [gpt-5-mini]} & \checkmark  & \checkmark & 2.4 & 33 & 30 & 59 & \phantom{0}4 & 37 & 66.0\% \\
\texttt{few-shot [gpt-5]} & \checkmark  & \checkmark & 2.5 & 36 & 27 & 54 & \phantom{0}9 & 36 & 66.7\% \\
\midrule
\texttt{pocketflow-agent} & \checkmark  & \checkmark & 2.2 & 38 & 25 & 32 & 31 & 56 & 57.6\%\\
\texttt{swe-agent}  & \checkmark  & \checkmark & 2.2 & 35 & 28 & 54 & \phantom{0}9 & 37 & 65.4\% \\
\midrule
\texttt{prismaDV [gemini-2.5-flash]} &  \checkmark & \checkmark & 2.1 & 59 & \phantom{0}4 & 40 & 23 & 27 & 81.4\%\\
\texttt{prismaDV [gemini-2.5-pro]} &  \checkmark & \checkmark & 1.7 & 59 & \phantom{0}4 & 42 & 21 & 25 & 82.5\%\\
\texttt{prismaDV [gpt-4.1]}& \checkmark & \checkmark & 2.1 & 56 & \phantom{0}6 & 38 & 25 & 31 & 77.8\%\\
\texttt{prismaDV [gpt-5-mini]}& \checkmark & \checkmark & 2.5 & 60 & \phantom{0}3 & 46 & 17 & \underline{20} & \underline{85.7\%} \\
\texttt{prismaDV [gpt-5]}& \checkmark & \checkmark  & 2.6 & 61 & \phantom{0}2 & 48 & 15 & \textbf{17} & \textbf{87.8\%}\\
\bottomrule
\end{tabular}}
\caption{Results for constraint discovery from data-code pairs in \icdbench{}, with outlier detection methods, task-agnostic data unit test generation, LLM-based prompting and software engineering agents as baselines. The best values are marked in bold, the second best values are underlined. \system{} outperforms all baselines by a large margin of more than 20 points in F1 score.}
\label{tab:constraint-discovery}
\end{table*}

We evaluate the ability of our system to discover individual constraints from data-code pairs, which is the foundation for generating high-quality data unit tests.

\header{Experimental setup and baselines} We leverage \icdbench{}, our individual constraint discovery benchmark, introduced in~\Cref{sec:benchmarks-icdb}. Our evaluation protocol is as follows. For each of the 63 cases in the benchmark, we first expose the data-code pair in the form of a passing data sample and the example code to the method to evaluate. Next, we expose the data-to-pass and the data-to-reject from each case to the method and ask it to decide whether the data is valid. This leads to 126 binary decisions per method to evaluate, for which we compute the F1 score as quality metric. 
We evaluate \system{} with different LLMs from OpenAI and Google as well as a large range of additional baselines:

\begin{itemize}[leftmargin=*]
\item \textit{Outlier detection} -- We evaluate classic ML methods for outlier detection such as an isolation forest~\cite{liu2008isolation} (refered to as \texttt{isolation-forest}) and a one-class SVM~\cite{scholkopf2001estimating} (\texttt{one-class-svm}). In addition, we evaluate an adaption of the ``partition summarisation'' approach proposed in~\cite{redyuk2021automating} (to which we refer as \texttt{stats-novelty}), where we compute the proposed descriptive statistics on the data, and decide upon rejection via a Maximum Mean Discrepancy-based test~\cite{gretton2012kernel}. Note that these methods are not task-aware and make their decisions based on the data alone. For each case to evaluate, we fit the outlier detection model (with default parameters) on the data sample and subsequently ask it to detect outliers in the data-to-pass and the data-to-reject. We mark either of them as invalid if the model says that outliers are present. We cannot include \texttt{autotest}~\cite{chen2025auto} in our evaluation, as we did not manage to get its code to run despite several hours of trying.
\item \textit{Task-agnostic data unit test generation} -- We evaluate task-agnostic methods for data unit test generation, which generate their data unit test solely based on the data sample for each case to evaluate. In particular, we evaluate Deequ~\cite{schelter2018automating} (\texttt{deequ}) via its ``constraint suggestion'' feature~\cite{deequdoc}, and Tensorflow Data Validation~\cite{polyzotis2019data} (\texttt{tensorflow-dv}) via its ``schema inference'' feature. We apply the generated data unit on the data-to-sample and data-to-pass to make decisions for the benchmark. Note that we cannot include Great Expectations as a baseline, as it currently does not provide an automated way to generate data unit tests.
\item \textit{In-context learning with LLMs} -- We design two task-aware baselines, which use a single LLM call with a custom prompt to generate a data unit test in \texttt{pydeequ} syntax. The first baseline, referred to as \texttt{zero-shot} uses zero-shot prompt to an LLM which contains the data sample, the downstream code, the target column and asks for the list of constraints as result. The second baseline \texttt{few-shot} extends this prompt with two manually crafted few-shot examples. We evaluate both baselines with various LLMs from OpenAI and Google.
\item \textit{Agentic systems} -- We treat data unit generation as a software engineering task and ask two agentic systems for software engineering to generate a data unit test in \texttt{pydeequ} syntax, based on the data sample and code from the benchmark, as well as a short task description. In particular, we evaluate the pocketflow agentic code generator~\cite{pocketflow} (\texttt{pocketflow-agent}), which is based on repeated series of code generation, testing and revision. We provide it with GPT-4.1 as base model, give it a budget of three full iterations, and manually restart it up to three times when it crashes. In addition, we evaluate a variant of the popular software engineering agent SWE Agent~\cite{yang2024swe} (referred to as \texttt{swe-agent}), which is designed to automatically fix GitHub issues. We use the ``swe-agent-mini'' implementation with GPT-5 as base model, give it a budget of \$0.50 per case, as well as a task description with two few-shot examples.
\end{itemize}

\begin{table*}[h!]
\centering
{\small
\setlength\extrarowheight{-1pt}
\begin{tabular}{l|cc|cccc|ccc}
 & &  &
\multicolumn{2}{c}{\textbf{Data to Pass}} & \multicolumn{2}{c}{\textbf{Data to Reject}} &
\multicolumn{3}{c}{\textbf{Metrics}} \\
\textbf{Method} & \textbf{Exec.}  & \textbf{Non-exec.} & 
Passed$\uparrow$ & False alarm$\downarrow$ & Rejected$\uparrow$ & Missed$\downarrow$ &
Precision$\uparrow$ & Recall$\uparrow$ & \textbf{F1 Score}$\uparrow$ \\ \toprule
\texttt{deequ} & 67.0 & \phantom{0}0.0 & 134 & 693 & 560 & 113 & \phantom{0}65.8\% & \phantom{0}14.4\% & \phantom{0}24.2\% \\
\texttt{tensorflow-dv} & - & - & 352 & 475 & 456 & 217 & \phantom{0}65.1\% & \phantom{0}42.7\% & \phantom{0}50.3\% \\
\midrule
\texttt{zero-shot [gemini-2.5-flash]} & 48.3 & 20.3 & \phantom{0}20 & 807 & 668 & \phantom{0}5 & \phantom{0}81.9\% & \phantom{0}2.4\% & \phantom{0}11.0\% \\
\texttt{zero-shot [gemini-2.5-pro]} & 12.4 & \phantom{0}2.8 & 203 & 624 & 595 & 78 & \phantom{0}65.0\% & \phantom{0}22.8\% & \phantom{0}31.0\% \\
\texttt{zero-shot [gpt-4.1]} & 26.1 & \phantom{0}4.6 & 100 & 727 & 594 & 79 & \phantom{0}51.0\% & \phantom{0}12.7\% & \phantom{0}18.7\% \\
\texttt{zero-shot [gpt-5-mini]} & 17.1 & \phantom{0}4.7 & 164 & 663 & 584 & 89 & \phantom{0}65.7\% & \phantom{0}20.0\% & \phantom{0}30.4\% \\
\texttt{zero-shot [gpt-5]} & 34.1 & \phantom{0}3.5 & \phantom{0}75 & 752 & 661 & 12 & \phantom{0}80.3\% & \phantom{0}\phantom{0}8.7\% & \phantom{0}31.9\% \\
\midrule
\texttt{few-shot [gemini-2.5-flash]} & 33.8 & \phantom{0}9.8 & 131 & 696 & 615 & \phantom{0}58 & \phantom{0}65.8\% & \phantom{0}15.2\% & \phantom{0}35.5\% \\
\texttt{few-shot [gemini-2.5-pro]} & \phantom{0}9.6 & \phantom{0}1.1 & 230 & 597 & 566 & 107 & \phantom{0}69.9\% & \phantom{0}26.8\% & \phantom{0}43.0\% \\
\texttt{few-shot [gpt-4.1]} & 23.2 & \phantom{0}3.6 & 152 & 675 & 613 & \phantom{0}60 & \phantom{0}66.8\% & \phantom{0}18.5\% & \phantom{0}40.4\% \\
\texttt{few-shot [gpt-5-mini]} & 18.8 & \phantom{0}3.0 & 215 & 612 & 570 & 103 & \phantom{0}67.1\% & \phantom{0}25.2\% & \phantom{0}51.3\% \\
\texttt{few-shot [gpt-5]} & 25.8 & \phantom{0}2.0 & 191 & 636 & 614 & \phantom{0}59 & \phantom{0}75.8\% & \phantom{0}21.4\% & \phantom{0}47.2\% \\
\midrule
\texttt{swe-agent [gpt-5]} & 24.1 & \phantom{0}1.7 & 179 & 648 & 622 & \phantom{0}51 & \phantom{0}78.6\% & \phantom{0}20.3\% & \phantom{0}47.2\% \\
\midrule
\texttt{prismaDV [gemini-2.5-flash]} & 36.4 & \phantom{0}0.0 & 551 & 276 & 533 & 140 & \phantom{0}81.7\% & \phantom{0}67.0\% & \phantom{0}72.9\% \\
\texttt{prismaDV [gemini-2.5-pro]} & 28.5 & \phantom{0}0.0 & 597 & 230 & 478 & 195 & \phantom{0}76.1\% & \phantom{0}73.0\% & \underline{\phantom{0}73.9\%} \\
\texttt{prismaDV [gpt-4.1]} & 34.1 & \phantom{0}0.0 & 484 & 343 & 551 & 122 & \phantom{0}81.6\% & \phantom{0}57.8\% & \phantom{0}67.0\% \\
\texttt{prismaDV [gpt-5-mini]} & 42.9 & \phantom{0}0.0 & 560 & 267 & 484 & 189 & \phantom{0}76.0\% & \phantom{0}68.9\% & \phantom{0}71.7\% \\
\texttt{prismaDV [gpt-5]} & 44.5 & \phantom{0}0.0 & 715 & 112 & 368 & 305 & \phantom{0}70.5\% & \phantom{0}86.6\% & \textbf{\phantom{0}77.4\%} \\
\bottomrule
\end{tabular}}
\caption{Detection performance with respect to the impact of data errors on downstream tasks in \eidbench{}. The best scores are highlighted in bold, the second-best underlined. \system{} outperforms all baselines by a large margin. Exec.\ reports the average number of executable constraints, and Non-exec.\ reports the average number of non-executable constraints.}
\label{tab:error-impact}
\end{table*}

\header{Results and discussion} We list the results for this experiment in~\Cref{tab:constraint-discovery}. For methods that generate data unit tests in the form of \texttt{pydeequ} constraints, we detail the average number of constraints generated. Furthermore, we count how often a method correctly identified the data to pass (Data to Pass > Passed) and how often it produced a false alarm by flagging this data as invalid (Data to Pass > False alarm). Analogously, we count how often each method correctly flagged data to reject as invalid (Data to Reject > Rejected) and how often it missed the rejection (Data to Reject > Missed). We compute the F1-score from these counts as the final quality metric.

A first observation is that the outlier detection based approaches exhibit an extremely low performance. This is a validation of our benchmark design, since it shows that inspecting data alone is not sufficient to make correct decisions, but that an ``understanding'' of the code is required. The task-agnostic data unit tests, LLM prompting approaches and software engineering agents provide a mixed performance with the largest F1 scores in the high sixties (\texttt{deequ} with a score of 61.5\%, \texttt{few-shot [gpt-5]} with a score of 66.7\%, and \texttt{swe-agent} with a score of 65.4\%). We observe that prompting approaches and agentic systems often generate the correct ground truth constraint, but suffer from the fact that they generate additional constraints which do not hold on the data. Our system \system{} manages to outperform them all by a large margin of more than 20 points with an F1-score 87.8\% for \texttt{prismaDV [gpt-5]}. Furthermore, it provides the lowest number of mispredictions, which is an important metric for operational deployments, where a misprediction might lead to a data engineer having to inspect the data manually to no avail. In summary, these findings confirm that constraint discovery from code is a difficult task, which benefits from a dedicated system and cannot be sufficiently solved with prompting or general engineering agents alone. Beyond the F1 margin, \system{}'s step-wise decomposition makes each stage easier to inspect.

\subsection{End-To-End Error Impact}
\label{sec:exp-end-to-end}

Next, we evaluate the ability of \system{} to detect the impact of data errors on downstream task behavior end-to-end, a setting that directly reflects the practical value of task-aware data unit tests.

\header{Experimental setup} We leverage \eidbench{}, our end-to-end benchmark introduced in~\Cref{sec:benchmarks-eidb}. Our evaluation protocol is as follows. We evaluate on the five datasets in the benchmark, which contain 60 downstream tasks in total. For each task, we provide $D_\text{sample}$ and the ``assertion-stripped'' task code to the method to generate a data unit test. We then evaluate this test on the 25 corresponding data batches and compare pass/reject decisions to the ground-truth labels computed by \eidbench{} (\Cref{sec:benchmarks-eidb}), yielding 1,500 (task, batch) decisions. Note that while each batch $D_i$ is corrupted, it is not necessarily erroneous for every task; thus, the evaluation set contains both safe and erroneous cases (827 safe, 673 erroneous).
We evaluate \system{} with a selection of the baselines used in \Cref{sec:exp-icdb}. We omit anomaly detection approaches, the pocketflow agent and smaller LLM variants due to consistently poor performance in \icdbench{} from the the previous \Cref{sec:exp-icdb}.

\header{Results and discussion} We list the results for this experiment in \Cref{tab:error-impact}. For methods that generate data unit tests in the form of \texttt{pydeequ} constraints, we report the average number of executable constraints. We also count the non-executable constraints, which we exclude from the pass/reject decision (i.e., non-executable constraints do not cause rejection). We then report how often a method correctly marks a safe data partition as pass (Data to Pass > Passed) and how often it raises a false alarm on safe data (Data to Pass > False alarm). Similarly, we report how often erroneous data is correctly rejected (Data to Reject > Rejected) and how often it is missed (Data to Reject > Missed). We compute the F1 score from these counts as the final metric.

Zero-shot and few-shot baselines perform substantially worse than \system{}. They show a strong tendency to generate more false alarms than misses, which is reflected in the gap between precision and recall. This suggests that these models generate overly strict constraints. With our post-processing, all constraints generated by \system{} in the testing stage are executable. Across different backend LLMs, \system{} achieves consistent performance, with \texttt{prismaDV [gpt-5]} reaching the highest F1 score of 77.4\%. At the same time, different backbone LLMs exhibit different trade-offs between missed rejections and false alarms. For example, \texttt{prismaDV [gpt-5]} raises relatively few false alarms (112), but it misses many erroneous cases (305). In contrast, \texttt{prismaDV [gemini-2.5-pro]} misses very few erroneous cases (195), but it raises more false alarms (230). This highlights that \system{}'s preference depends on the backbone LLM, and practitioners can choose a model based on their tolerance for false alarms versus missed rejections.

\subsection{Optimizing \system{} with SIFTA}
\label{sec:exp-optimization}

In the following, we validate that our proposed prompt optimization approach SIFTA (\Cref{sec:approach_optimization}) improves \system{}'s performance.

\begin{table}[t]
    \centering
    {\small
    \setlength{\tabcolsep}{5pt}
        \begin{tabular}{ll|cccc}
            \textbf{Target} & & \multicolumn{4}{c}{\textbf{F1 Score per Prompt Type}} \\
            \textbf{Scenario} & \textbf{Dataset}       & \texttt{GEPA[P]} & \texttt{GEPA[S]} & \texttt{Manual}              & \texttt{SIFTA}            \\
            \toprule

            \multirow{7}{*}{\centering\rotatebox{90}{New Data}}
            & \texttt{students}      & 70.50\%        & 63.44\%      & 70.10\%             & 86.88\%          \\
            & \texttt{hr\_analytics} & 71.54\%        & 68.44\%      & 69.00\%             & 70.69\%          \\
            & \texttt{sleep\_health} & 54.85\%        & 69.63\%      & 58.42\%             & 71.30\%          \\
            & \texttt{IPL}           & 58.17\%        & 62.50\%      & 68.55\%             & 64.27\%          \\
            & \texttt{imdb}          & 62.28\%        & 68.08\%      & 64.06\%             & 68.60\%          \\
            \cmidrule(lr){2-6}
            & \textit{Mean}          & 65.34\%        & 66.42\%      & \underline{67.71\%} & \textbf{72.84\%} \\
            \bottomrule

            \multirow{7}{*}{\centering\rotatebox{90}{New Tasks}}
            & \texttt{students}      & 76.07\%        & 76.96\%      & 75.66\%             & 85.30\%          \\
            & \texttt{hr\_analytics} & 84.17\%        & 82.84\%      & 83.19\%             & 82.37\%          \\
            & \texttt{sleep\_health} & 52.45\%        & 51.54\%      & 58.70\%             & 56.98\%          \\
            & \texttt{IPL}           & 43.50\%        & 58.22\%      & 55.53\%             & 57.03\%          \\
            & \texttt{imdb}          & 48.01\%        & 49.58\%      & 56.43\%             & 54.75\%          \\
            \cmidrule(lr){2-6}
            & \textit{Mean}          & 64.81\%        & 63.83\%      & \underline{67.29\%} & \textbf{69.76\%} \\
            \bottomrule

            \multirow{7}{*}{\centering\rotatebox{90}{\makecell{New Data \\+ New Tasks}}}
            & \texttt{students}      & 73.33\%        & 57.34\%      & 65.85\%             & 75.68\%          \\
            & \texttt{hr\_analytics} & 76.76\%        & 78.37\%      & 75.62\%             & 78.92\%          \\
            & \texttt{sleep\_health} & 61.31\%        & 60.40\%      & 68.33\%             & 65.55\%          \\
            & \texttt{IPL}           & 46.88\%        & 58.97\%      & 56.44\%             & 60.76\%          \\
            & \texttt{imdb}          & 51.58\%        & 55.45\%      & 64.81\%             & 55.41\%          \\
            \cmidrule(lr){2-6}
            & \textit{Mean}          & 64.45\%        & 62.11\%      & \underline{66.11\%} & \textbf{67.75\%} \\
            \bottomrule

        \end{tabular}}
    \caption{Impact of optimizing the prompts of \system{} for different scenarios. \texttt{SIFTA} outperforms all baselines on average across all three test scenarios, with the largest average gain in the \emph{New Data} scenario.}
    \label{tab:averaged-optimization-results}
\end{table}

\begin{table}[h!]
\centering
{\small
\begin{tabular}{l|c|c}
\textbf{Variant} & \textbf{F1 score on \eidbench{}} & \textbf{Delta}\\
\toprule
\texttt{prismaDV [gpt-5]}     & 77.35\% & - \\
\midrule
w/o multicolumn constraints & 76.43\%& -0.92\% \\
w/o dataflow analysis & 76.87\% & -0.48\% \\
w/o assumption inference & 77.18\% & -0.17\% \\
\bottomrule
\end{tabular}}
\caption{Ablation results on \eidbench{} for different variants of \system{} with GPT-5 as backing model. Each system module contributes to the overall performance.}
\label{tab:ablation}
\end{table}

\header{Experimental setup} We evaluate prompt optimization for \system{} on \eidbench{} under the three generalization targets defined in \Cref{sec:optimization_problem}: \textit{New Data}, \textit{New Tasks}, and \textit{New Data + New Tasks}. We follow the end-to-end pass/reject protocol from \Cref{sec:exp-end-to-end} and report F1 scores. For each dataset $D$, we split its downstream tasks into $T_{\mathrm{train}}$, $T_{\mathrm{eval}}$, and $T_{\mathrm{test}}$ with a 3:3:4 ratio.
We also split the corresponding 25 data batches into an observed set $D_{\mathrm{obs}}$ and a held-out set $D_{\mathrm{new}}$ with a 1:1 ratio. We optimize prompts on $T_{\mathrm{train}}\times D_{\mathrm{obs}}$ and use $T_{\mathrm{eval}}\times D_{\mathrm{obs}}$ for evaluation; we then test on three target scenarios: \textit{New Data}: $(T_{\mathrm{train}} \cup T_{\mathrm{eval}})\times D_{\mathrm{new}}$,  \textit{New Tasks}: $T_{\mathrm{test}}\times D_{\mathrm{obs}}$, and \textit{New Data + New Tasks}: $T_{\mathrm{test}}\times D_{\mathrm{new}}$.

We compare against the following baselines: \texttt{Manual}, our hand-tuned set of prompts for \system{} used in Section~\ref{sec:exp-end-to-end}; and GEPA~\cite{agrawal2025gepa}, a general-purpose prompt optimizer that learns from the Pareto frontier of optimization attempts. We include two GEPA variants. First, for \texttt{GEPA[S]}, we configure GEPA following principles aligned with the original paper. For each training or evaluation example, we provide GEPA with a single simple prompt that takes the task and column profile information as input; after obtaining the output, we compute the F1 score between the prediction and the ground-truth safeness of $D_{\mathrm{new}}$ on the task. Second, for \texttt{GEPA[P]}, we use GEPA to optimize \system{} modules, where each prompt describes a module's function in one sentence. The same set of basic prompts is optimized by \texttt{SIFTA}. For all optimization approaches, we apply early stopping and terminate if the proposer fails to outperform the current best prompts for 20 proposing iterations.

\texttt{GEPA[P]}, \texttt{GEPA[S]}, and \texttt{SIFTA} are all assigned a budget of 15 full evaluation executions, which is more than enough for the optimization process to converge. For \texttt{SIFTA}, we recalculate the training set condensation stage every 5 full evaluation executions. We repeat the experiment with two different random seeds, and set the training sample batch size to three for both methods. We use \texttt{gpt-5} as the prompt proposer and \texttt{gpt-4.1-mini} as the backbone LLM of the data validation system for both methods.

\header{Results and discussion}
We list the optimization performance in~\Cref{tab:averaged-optimization-results}; for each method, we run the optimization twice and report the averaged results. \texttt{SIFTA} outperforms \texttt{GEPA[P]}, \texttt{GEPA[S]} and the \texttt{Manual} baseline on average across all three target scenarios. In the \emph{New Data} scenario, \texttt{SIFTA} achieves the largest average gain, exceeding the GEPA variants by 5.13 points. In the more challenging scenarios that require generalizing to new tasks, \texttt{SIFTA}'s optimization is slightly better than the \texttt{Manual} baseline. By contrast, the GEPA variants do not consistently match the performance of the \texttt{Manual} baseline across all three scenarios. This aligns with our observation that \texttt{GEPA[P]} struggles to propose effective updates for \system{}'s coupled, multi-module prompts. This indicate that GEPA cannot optimize PrsimaDV when the feedback text didn't contains enough information like SIFTA. \texttt{GEPA[S]} can improve the single initial prompt but the optimization result can not match \texttt{SIFTA}.


\subsection{Ablation Study}
\label{sec:exp-ablation}

Next, we conduct an ablation study to systematically validate that each system component introduced in \Cref{sec:approach} contributes meaningfully to the overall performance of \system{}. Specifically, we manually disable individual system modules and API methods in \texttt{prismadv [gpt-5]} and evaluate each resulting system variant on \eidbench{}, reporting the mean F1 score averaged over all five datasets included in the benchmark. The results shown in \Cref{tab:ablation} confirm that every component plays a beneficial role in the system, as disabling any single component consistently leads to a measurable decrease in F1 score.


\section{Conclusion}

This paper introduced \system{}, a task-aware data validation approach that synthesizes specialized data unit tests based on both the data and downstream tasks. To improve \system{}’s performance on specific datasets, we further proposed SIFTA, an optimizer that efficiently adapts the prompts used in \system{}’s modules based on failure precision on training examples. To evaluate task-aware data validation systems from diverse perspectives, we introduced two novel benchmarks, \icdbench{} and \eidbench{}.

\header{Limitations and Future Work} Our prototype focuses on single-file tasks over a single table. Supporting multi-file code and multi-table inputs will require additional engineering, such as tracking dataflow across scripts and reasoning about joins and derived tables, but remains conceptually compatible with task-aware assumption inference. A promising next step is to extend our benchmarks with real industry workloads, enabling evaluation under more realistic codebases, data distributions, and operational constraints. By treating textual data assumptions as an intermediate representation, \system{} has the potential to support multiple validation DSLs, such as Great Expectations and Python \texttt{assert} statements. SIFTA currently relies on training batches with observed errors, which can be difficult to obtain in practice. Automating error injection could synthetically generate training batches, enabling optimization without relying on naturally occurring failures and making \system{} more suitable for cold-start settings. In SIFTA, we adopted a simple greedy search strategy, we leave evaluating more advanced strategies (e.g., Monte Carlo tree search) for future work.

\clearpage
\bibliographystyle{ACM-Reference-Format}
\bibliography{reference}

\end{document}